%% file: main.tex
\documentclass[11pt,twoside,twocolumn,a4paper]{article}

\usepackage{cvww}
\usepackage{times}
\usepackage{epsfig}
\usepackage{graphicx}
\usepackage{amsmath}
\usepackage{amssymb}

\input{style/packages.tex}

\cvwwfinalcopy % *** Uncomment this line for the final submission

\def\cvwwPaperID{27} % *** Enter the CVWW Paper ID here

\ifcvwwfinal\fi

\def\mytitle{Normalization of Neural Networks using Analytic Variance Propagation}

\begin{document}

%%%%%%%%% TITLE
\title{\mytitle}
\author{Alexander Shekhovtsov, Boris Flach\\
Department of Cybernectics, Czech Technical University in Prague\\
Zikova 4, 166 36 Prague\\
{\tt\small shekhovt@cmp.felk.cvut.cz}
}
\maketitle
\ifcvwwfinal\thispagestyle{fancy}\fi

%%%%%%%%% ABSTRACT
\begin{abstract}
We address the problem of estimating statistics of hidden units in a neural network using a method of analytic moment propagation. These statistics are useful for approximate whitening of the inputs in front of saturating non-linearities such as a sigmoid function. This is important for initialization of training and for reducing the accumulated scale and bias dependencies (compensating covariate shift), which presumably eases the learning. In batch normalization, which is currently a very widely applied technique, sample estimates of statistics of hidden units over a batch are used. The proposed estimation uses an analytic propagation of mean and variance of the training set through the network. The result depends on the network structure and its current weights but not on the specific batch input. The estimates are suitable for initialization and normalization, efficient to compute and independent of the batch size. The experimental verification well supports these claims. However, the method does not share the generalization properties of BN, to which our experiments give some additional insight.
\end{abstract}

%%%%%%%%% BODY TEXT

\section{Introduction}

{\em Batch normalization} (BN)~\cite{IoffeS15} is a widely applied method which is known to improve learning speed and performance of difficult networks. It is based on a whitening normalization that requires computing the mean and variance statistics of all activations over the training set. In~\cite{IoffeS15} these statistics are approximated by those over a batch. %Other approximations are possible. For example, {\em layer normalization}~\cite{Ba-2016-Layer-Norm} uses statistics of different units within the same layer.

Since~\cite{IoffeS15}, there have appeared a number of different normalization methods. A good overview is given in~\cite{Gitman-17} who categorize current methods in three groups: methods based on sample statistics over different groups of hidden units~\cite{Ba-2016-Layer-Norm, Ulyanov-16, Ren-2016a}, modifications of BN~\cite{Salimans-16,HofferHS17,Ioffe17,LiaoKP16} and methods normalizing weights instead of activations~\cite{Salimans2016WeightNA,ArpitZKG16,xiang2017effects}.

The proposed technique is a follow-up application of the work~\cite{Shekhovtsov-17}, where a feed-forward propagation of uncertainties (variances) in neural networks (NNs) is proposed, making NNs and Bayesian networks more alike. %Training such variance propagating network with a normalization based on the same technique was demonstrated. 
In this work we apply the idea of variance propagation to analyze standard networks. Namely, we are interested in estimating means and variances of activations in a given network provided basic statistics of the dataset.
We show that the method~\cite{Shekhovtsov-17} is suitable for this task, can be implemented very efficiently for CNNs and conduct a detailed experimental study.

The method is also related to {\em normalization propagation}~\cite{ArpitZKG16} and {\em deep information propagation}~\cite{Schoenholz2016DeepIP} as will be discussed in \cref{sec:related}.
% and {\em fast dropout training}~\cite{wang2013fast}
%Our model can instead compute an analytic estimate of such statistics in one pass. 

\subsection{Background}
\paragraph{Normalization Methods}
\citet{IoffeS15} proposed the following transformation to be applied in-front of saturating non-linearities (such as logistic sigmoid):
\begin{equation}\label{BN}
X' = \frac{X - \mu}{\sigma} s + b,
\end{equation}
where $s$ and $b$ are new parameters.
Its purpose is twofold: 1) to make sure that the non-linearity {\em on average} receives a signal in a range where the output is not saturated and 2) to make the average scale and bias of activations controlled by local parameters $s$ and $b$ rather than by the cumulative effect of many layers. %It was proposed~\cite{IoffeS15} the transform 

The ideal whitening is achieved when $\mu =\E[X]$ and $\sigma^2 = \Var[X]$ are the expectation statistics of $X$ over the dataset. While it is too costly to compute these expectations accurately for all hidden units in a deep network, they can be approximated in several ways. %These statistics are however intractable.
BN estimates $(\mu, \sigma^2)$ as sample statistics over a batch. Other estimates / approximations are possible. For example, layer normalization~\cite{Ba-2016-Layer-Norm} uses sample statistics of different units across the spatial dimension of a layer.

{\em Weight Normalization} (WN)~\cite{Salimans2016WeightNA}, applied to the input $X=w\T Z$ can be viewed as setting $\mu=0$ and $\sigma = \|w\|$ in~\eqref{BN}. This choice indeed matches the expectations of $X$ when $Z$ has zero mean and unit variance~\cite{Salimans2016WeightNA}.
Substituting, we obtain the formula $X' = w\T Z / \|w\| \cdot s + b$, which just normalizes the weight locally. The resulting variable $X'$ as well as a non-linear transform of it, $f(X')$, do not longer have a zero mean and unit variance (unless it is assumed that $s=1$ and $b=0$) and the chain of arguments breaks down. Despite such simplicity, WN has a positive effect on learning~\cite{Salimans2016WeightNA,Gitman-17}.
%The assumption on $Z$ however breaks down after a non-linearity.
These two techniques will be the baseline methods in our experiments.
\paragraph{Invariances}
Let $X$ be the output of a linear transform: $X = w\T Z + a$. Then $X'$ in~\eqref{BN} is invariant \wrt the bias $a$ and the scale of weights $\| w \|$. This holds true for the exact expectations $\mu$, $\sigma^2$ as well as for their estimates used by weight normalization, batch normalization and our method. The invariance to the bias is evident, it holds as soon as the estimate of $\mu$ satisfies the linearity of the expectation: $\E [X] = w\T \E [Z] + a$.
The scale invariance follows from that the variance and its used estimates are 2-homogenous: $\Var [\alpha w\T Z]$ = $\alpha^2 \Var [w\T Z]$. In this case there holds: %normalization~\eqref{BN} satisfies
\begin{align}
\frac{\E [\alpha w\T Z]}{\sqrt{\Var[\alpha w\T Z]}} = \frac{\E [w\T Z]}{\sqrt{\Var[w\T Z]}},
\end{align}
so the scale $||w||$ does not matter.
For BN, the sample variance of $w\T Z$ expresses as $\< w w\T, C\>_F$, where $C = \frac{1}{n-1}\sum_i z_i z_i\T$ is the sample covariance matrix over a batch $\{z_i\}_i$ and $\<\cdot, \cdot\>_F$ is the Frobenius inner product. This expression is 2-homogenous in $w$.
\par
With these invariances we see that the new parameters $s,b$ in the expression~\eqref{BN} introduce back exactly the same number of degrees of freedom that are projected-out by the normalization. The difference is that the overall scale and bias are controlled now by the local parameters $s, b$ rather than by the cumulative effect of the preceding layers.
%
%
%For x' = \frac{w\T (z - M)}{\sqrt{\< w w\T, C\>_F}} s + b,
%is less straightforward for BN. 
%Suppose that $X$ was a result of a linear transform $X = w\T Z + a$. Let $M$ and $C$ denote the sample mean and sample covariance of $Z$ over a batch. The normalized form~\eqref{BN} can be expressed as
%The normalized form~\eqref{BN} using equations~\eqref{linear} can be written as
%\begin{align}\label{BN-inv}
%x' = \frac{w\T (z - M)}{\sqrt{\< w w\T, C\>_F}} s + b,
%\end{align}
%where $\<\cdot, \cdot\>_F$ denotes the Frobenius inner product. It is clear that this expression is invariant to the initial bias $a$ and to the scale $\|w\|$.
%It is obvious that~\eqref{BN} is invariant to the initial bias $a$ as soon as the estimate of $\mu$ is linear, as the true expectation is. 
%Less obvious is that, \eg, BN is also invariant to the change of scale $\|w\|$ (which we show below). 

\paragraph{Initialization}
There are two ways to introduce a normalization: preserving the equivalence with the original network and resetting the scale and bias.
In the first case, the new parameters $s$, $b$ in~\eqref{BN} are initialized as $s = \sigma$ and $b = \mu$, which makes~\eqref{BN} an identity.
The network performance is preserved. This does not help to achieve a non-saturating regime discussed above but can be useful for further optimization of an already pretrained network.

In the second case, $s$ and $b$ are initialized, \eg, as $s = 1$ and $b = 0$. %This efficiently initializes the network to a point where non-linearities are not saturated on average.
%New variables $s, b$ introduce back exactly the same number of degrees of freedom that were projected-out, but they are now reset to the new values.
%
%Suppose that $X$ was a result of a linear transform $X = w\T X + a$.
%The normalized form~\eqref{BN} using equations~\eqref{linear} can be written as %becomes invariant to $w_0$ and furthermore it becomes invariant to $$
%\begin{align}
%X' = \frac{w\T (Z - \E[Z])}{\sqrt{\<w w\T, \Cov[Z]\>_F}} s + b,
%\end{align}
%where $\<\cdot, \cdot\>_F$ denotes the Frobenius inner product. It is clear that this expression is invariant to the initial bias $a$ and to the scale $\|w\|$. 
%
%The later method can be applied to initialize network parameters~\cite{Salimans2016WeightNA}. 
%
Starting from some initialization of network weights, such as random or orthogonal, this projecting initialization approximately compensates accumulated biases and scaling of linear and non-linear transforms and efficiently reinitializes the network to a point where non-linearities are not saturated on average. It makes the network training invariant to a scale-bias preprocessing transformation of the input images and the initial scale of the random weights. Converting the model back to the equivalent unnormalized form will preserve these properties and result in a good initialization point (\eg,~\cite{Salimans2016WeightNA}). %Such initialization makes sure that we are not starting in a point where a non-linearity is saturated for all inputs because of the bias or where the gradients are exploding. 
%The normal operation of BN~\cite{IoffeS15} initializes with $s$ random and $b=0$.
%
%
\paragraph{Shortcomings of BN}
Let us summarize the known issues / challenges with using BN in practice. The testing requires an up-to-date whole dataset statistics. This makes it more difficult to keep track of the training/validation error during the training.
The application to recurrent networks is problematic: one has to unfold running statistics over recurrences. The computation overhead is not negligible: up to 25\% according to~\cite{Gitman-17}. The behavior is more stochastic than other methods~\cite{Salimans2016WeightNA} and destabilizing in GAN, where weight normalization performs better~\cite{xiang2017effects}. The batch size is a sensitive parameter affecting convergence and generalization properties~\cite[Fig. 7.38]{Schilling-16}.

\section{Proposed Technique}
The proposed technique uses the same normalized form~\eqref{BN}, but the mean and variance are estimated analytically rather than from a batch sample. The analytic method~\cite{Shekhovtsov-17} is as follows.
A basic neural network composes linear transforms and coordinate-wise non-linearities. Let us assume that $X$ is a vector-valued r.v. with components $X_i$ having statistics $\mu_i$ and $\sigma^2_i$.
The statistics of a linear transform $Y= w\T X$ are given by
\begin{subequations}\label{linear}
\begin{align}
\mu' &= \E[Y] = w\T \E[X] = w\T \mu,\\
\sigma'^2 & = \sum_{ij} w_i w_j \Cov[X] \approx \sum_{i} w_i^2 \sigma_i^2,
\end{align}
\end{subequations}
where $\Cov[X]$ is the covariance matrix of $X$. The approximation of the covariance matrix by its diagonal is exact when $X_i$ are uncorrelated. The correlation of the outputs depends on the current weights $w$ and is zero when the weights are orthogonal. Note that random vectors are approximately orthogonal and therefore the assumption is plausible at least on initialization.

\begin{figure}
\centering
\includegraphics[width=\linewidth]{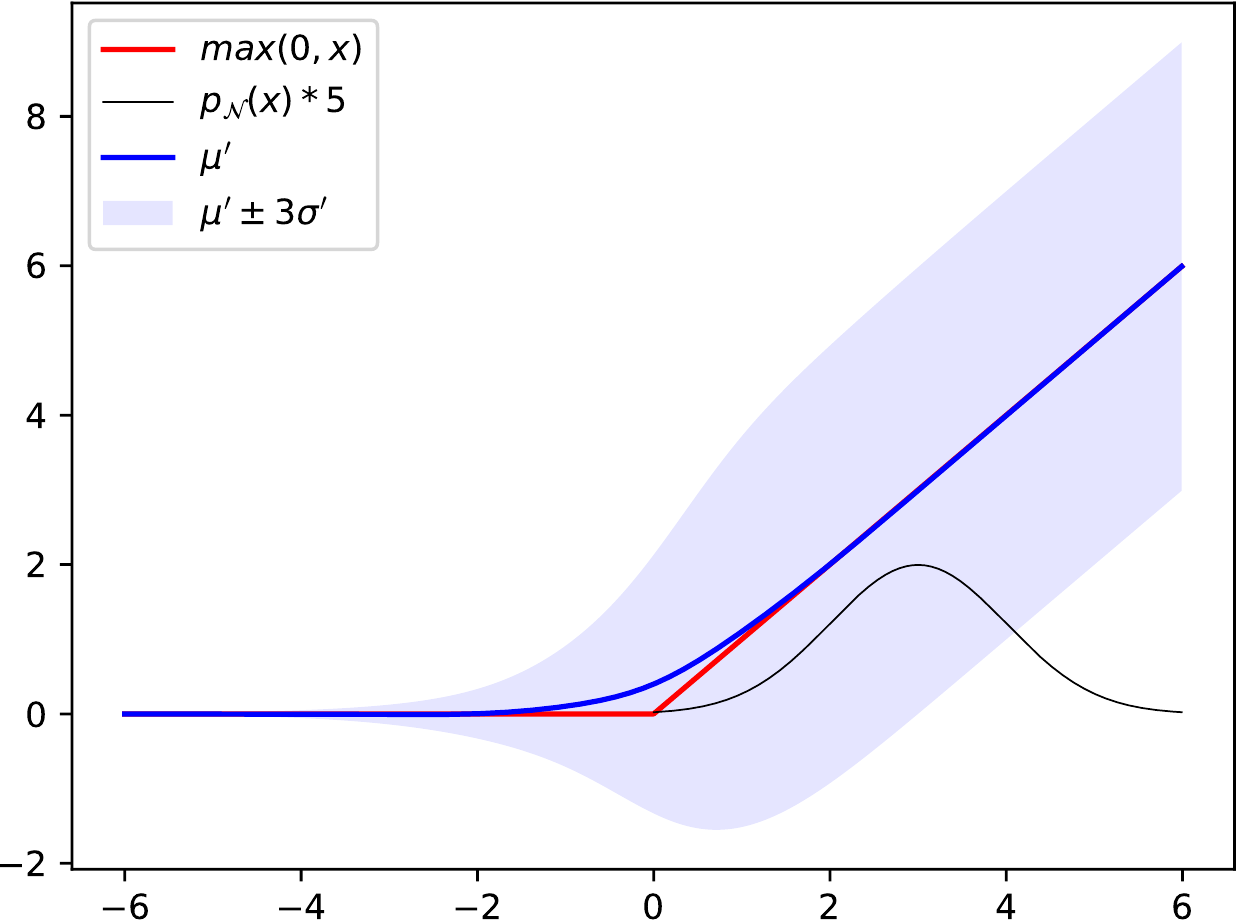}
\caption{\label{fig:relu}
Propagation of uncertainty through ReLU (red). The black curve shows an example of input $X$: a normal pdf with mean $\mu=3$, $\sigma=1$ and support $\mu\pm3\sigma$, amplified in value for visibility. The expectation of ReLU \vs the mean $\mu$ is shown as a blue curve. The std $\sigma'$ of ReLU is illustrated via the set $\mu'\pm 3\sigma'$ shown as shaded area. When the input $\sigma$ is different from $1$ the plot in coordinates $\mu'/\sigma$ vs. $\mu/\sigma$ stays the same.
}
\end{figure}

In order to treat non-linearities, let $X$ be a scalar \rv with statistics $\mu, \sigma^2$ and $Y=f(X)$. One general method of variance estimation is to linearize $f$ around $\mu$ and apply the propagation in the linear case. However, this is only applicable to small variances, such that the linear approximation holds for all likely values of $X$. When we deal with statistics over a dataset, the variances should be assumed large.
A more suitable approximation used in~\cite{Shekhovtsov-17} computes expectations
\begin{subequations}\label{integrals}
\begin{align}
\mu' &= \E[f(X)] = \int f(x) p(x) dx,\\
\sigma'^2 &= \int f^2(x) p(x) dx - \E[f(X)]^2,
\end{align}
\end{subequations}
assuming that $X$ is normally distributed: $X \sim \N(\mu, \sigma^2)$. This assumption is plausible when $X$ is a linear combination $X = w\T Z + a$, in which $Z_i$ are independent and weights are random. Empirically, this is a good approximation in many cases in practice\footnote{There are variants of central limit theorem for non-i.i.d. 	and even weakly dependent variables \url{https://en.wikipedia.org/wiki/Central_limit_theorem}. See~\cite[Figs 2, 3]{wang2013fast} for experimental illustration of related assumptions in a network with dropout noise.}.
This method is not a general one as it requires to compute integrals analytically, but for many cases of practical interest it is tractable. In particular it is the case for sigmoid $f(x) = 1/(1+e^{-x})$ and ${\rm ReLU}(x)=\max(0,x)$ non-linearities and for $\max(x_1, x_2)$, which can be used to implement max-pooling and max-out, see~\cite[Table 2]{Shekhovtsov-17}. Let us illustrate the case for ReLU.
Both integrals~\eqref{integrals} can be taken (\eg, ~\cite[\parSym C.3]{Shekhovtsov-17}) and the result can be expressed as
\begin{subequations}\label{relu}
\begin{align}
\mu' &= \mu \Phi(a)+\sigma \phi(a)\\
\sigma'^2 &= \sigma^2 \R(a),
\end{align}
\end{subequations}
where $a= \mu/\sigma$, $\phi$ is the pdf of the standard normal distribution, $\Phi$ is its cdf and $\R(a) = a\phi(a)\,{+}\,(a^2{+}1)\Phi(a)\,{-}\,(a\Phi(a){+}\phi(a))^2$.
We see that both integrals express though functions of a single variable $a$. Even though they involve a non closed-form function $\Phi$, they can be accurately approximated. \cref{fig:relu} illustrates this solution.
\paragraph{Invariances}
It is straightforward to see that linear propagating equations~\eqref{linear} satisfy the properties that $\mu'$ is linear and $\sigma'^2$ is 2-homogenous and therefore the scale-bias invariance holds.

%The propagation equations~\eqref{linear},\eqref{relu} 
\subsection{Efficiency}
To compute statistics of all hidden units, the approximation can be applied layer-by-layer, propagating mean and variance.
Let us consider a deep NN where normalization~\eqref{BN} is applied. %If $\mu$, $\sigma$ are the exact statistics of $X$, the normalized quantity
The quantity
\begin{align}\label{normalized}
\frac{X_i-\mu_i}{\sigma_i},
\end{align}
where $\mu$, $\sigma^2$ are statistics of $X$, has zero mean and unit variance. % (consider applying~\eqref{linear} to evaluate statistics of~\eqref{normalized}).
% (this is so because the expression~\eqref{normalized} is linear in scalar $X_i$ and approximation~\eqref{linear} is exact in this case). 
It implies that the computation of statistics in a normalized network decouples as follows:
\begin{itemize}
\item Start from the dataset statistics $(\mu^{0}, \sigma^{0})$, which can be estimated prior to learning. They are commonly used for the initial data whitening transform. If such a whitening transform has been already applied we may assume $(\mu^{0}, \sigma^{0}) = (0,1)$. Propagate the moments until the first normalization layer. 
\item Propagate from one normalization layer to the next one assuming the input statistics are $(0,1)$.
\end{itemize}
%
%When the network is composed of blocks containing ${\tt linear}(W^k)$, ${\tt norm}(s^k,b^k)$ and ${\tt activation}$, the statistics for the normalization in block $k$ will depend only on parameters $W^k$ and $s^{k-1}$, $b^{k-1}$, but not on any other parameters or data.
The computation thus decouples in block of layers delimited by the normalization layers and the result of the normalization of such a block depends on the parameters only inside the block but not on any other parameters or data. Consider a network composed of blocks of the form $({\tt linear}(W^k)$, ${\tt norm}(s^k,b^k)$, ${\tt activation}$). Then the statistics for the normalization in block $k$ will depend only on parameters $W^k$, $s^{k-1}$, $b^{k-1}$.
\par
Further on, to compute the normalization in a convolutional NN we do not need to work with spatial dimensions. The normalized input statistics $(0,1)$ are the same over spatial dimensions and channels. The scale-bias parameters $s$, $b$ are the same over spatial dimensions but not over channels. A convolutional filter $W_{i,j,k}$ (per output channel) with $i$ running over input channels and $j,k$ over spatial dimensions, applied to a tensor which is constant across spatial dimensions, is equivalent to applying a linear transform $w$ with components $w_i = \sum_{j,k} W_{i,j,k}$ to a vector over channels. The propagation of the variance reduces similarly. The whole normalization is thus independent of the image and batch sizes and has a complexity similar to that of weight normalization.
%
%\paragraph{Invariances}
%Let $X$ be the result of a linear transform $X = w\T Z + a$. The normalized form~\eqref{BN} using propagation~\eqref{linear} can be written as
%\begin{align}
%X' = \frac{w\T (Z - \E[Z])}{\sqrt{\sum_{i}w_i^2 \Var[Z_i]}} s + b,
%\end{align}
%It is clear that this expression is invariant to the initial bias $a$ and to the scale $\|w\|$.

\paragraph{Initialization}
In our case, just the initial statistics $(\mu^0, \sigma^0)$ are needed for the initialization. The equivalence-preserving initialization is exact for the whole dataset. 
In the projecting initialization, approximate expectations $\E[X_i]$ and $\Var[X_i]$ depend on the parameters of all layers down to the preceding normalization layer as explained above. % and not only on the immediately preceeding $w$ as in WN.
%The projecting initialization approximately compensates accumulated biases and scaling of linear and non-linear transforms (note that $\E[Z]$ and $\Var[Z_i]$ depend on the parameters of all layers down to the preceding normalization).

\paragraph{Advantages}
We see the following advantages compared to~\cite{IoffeS15}. The proposed normalization has very little overhead for CNNs, also considering back-propagation.
There is no dependency on batch size.
The normalization is exactly the same during training and testing, we can easily convert between unnormalized and normalized forms while preserving the equivalence.
Because the normalization decouples, there is no limitation to apply it in recurrent networks. 
The normalization is continuously differentiable and is non-stochastic. 
If dropout is applied during training, the normalization takes it into account analytically (dropout is a multiplication by a Bernoulli \rv with known statistics, see details in~\cite{Shekhovtsov-17}).
%even when a stochastic regularization such as dropout~\cite{srivastava14a} is used (it is taken into account analytically).
%When a stochastic regularization is used in the network, such as dropout~\cite{srivastava14a}, the normalization computes expected values and remains smooth and non-stochastic.
Similar advantages hold for applying the method as initialization.
%
%
%
%The same is true for our approximation: if $\mu$ and $\sigma$ are approximated 
%
%In a deep NN, statistics after linear transforms can be applied in normalization~\eqref{BN}. Then, the linear transform becomes

%However, in order for this estimate to be accurate, it is necessary that the linear approximation is accurate not only locally at the mean but 

\section{Related Work}\label{sec:related}
Our method reduces to normalization propagation~\cite{ArpitZKG16} when parameters $s, b$ are not present. In this case the normalization results in centering of the ReLU non-linearity by a constant scale and bias as in~\cite{ArpitZKG16}: when input statistics are $(0,1)$ then so are the output statistics. A similar constant centering appears in self-normalizing networks~\cite{Klambauer-SELU}.
%When our method is applied with resetting $s,b$, the normalization reduces
Thus, as an initialization scheme, it is equivalent to~\cite{ArpitZKG16,Klambauer-SELU}. However, when the scale-bias parameters are enabled and varied, our method smoothly tracks the covariate shift, while~\cite{ArpitZKG16} does not. Disabling scale and bias parameters as in~\cite{Klambauer-SELU} defines a different class of models with fewer degrees of freedom. Since scale-bias parameters may be different per channel, our normalization depends as well on the current weights $w$. Further, note that applying dropout will change the statistics, which is taken into account in our method.
\par
The technique~\cite{Schoenholz2016DeepIP} performs analysis of neural networks using mean field theory, which is related to our variance propagation. It allows to estimate quantities such as mean activations and even norms of gradients assuming that all weights are random. While they focus on asymptotic scenarios \wrt depth and address a limited family of architectures, the results are highly relevant for initialization methods.

%The idea of analytical variance propagation is similar to self-normalizing networks~\cite{Klambauer-SELU}, where an analytical estimate of the mean and variance of an ELU unit is used to set the scaling and bias (once and for all) such that the output is normalized. They do not estimate it during training and need no normalization.
%
%A scheme proposing to compensate the bias of ReLU based on the preceding parameters $s^{k-1}$, $b^{k-1}$ was proposed by~\cite{ArpitZKG16} and termed normalization propagation. It has something in common with our propagation but does not take into account variances and considers only a special case. %They do not consider variance and 

Works~\cite{Gitman-17,Salimans2016WeightNA} discuss practical and theoretical properties of normalization schemes such as the effect of the reparametrization as a preconditioner, \etc.
%
%\subsection{Invariances of BN}
%Suppose that $X$ was a result of a linear transform $X = w\T Z + a$. Let $M$ and $C$ denote the sample mean and sample covariance of $Z$ over a batch. The normalized form~\eqref{BN} can be expressed as
%%The normalized form~\eqref{BN} using equations~\eqref{linear} can be written as
%\begin{align}\label{BN-inv}
%x' = \frac{w\T (z - M)}{\sqrt{\< w w\T, C\>_F}} s + b,
%\end{align}
%where $\<\cdot, \cdot\>_F$ denotes the Frobenius inner product. It is clear that this expression is invariant to the initial bias $a$ and to the scale $\|w\|$.
%
%
%
\section{How (Not) to Make a Comparison}
\begin{figure}
\includegraphics[width=\linewidth]{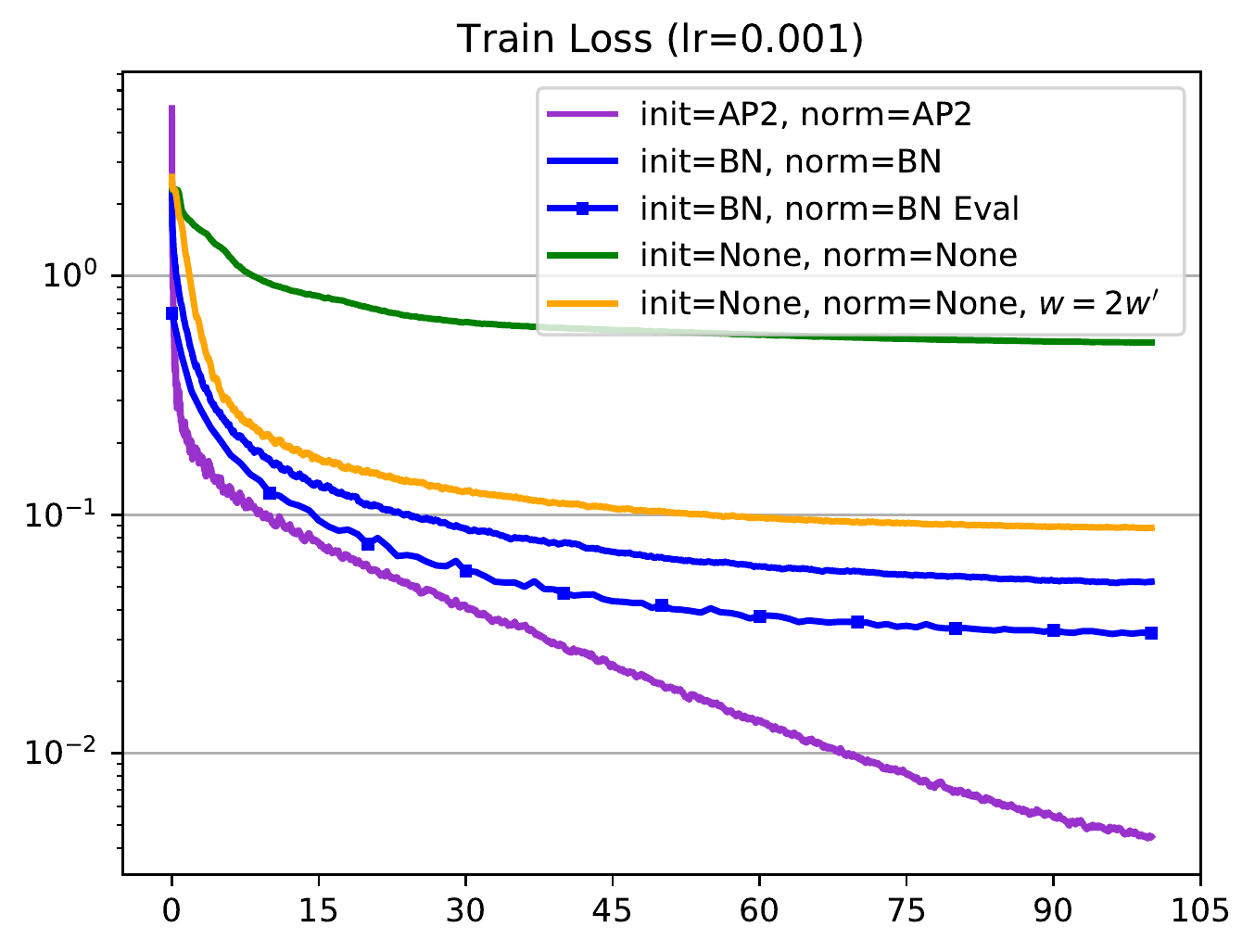}
\caption{\label{fig:naive} Comparison of normalization methods at equal learning rate using method-specific initializations. {\tt AP2} stands for the proposed method.  Solid curves show the running average of the loss (exponentially weighted). In this plot, the effect of the initialization is not clear. In addition, the speed of training highly depends on the parametrization: using a parametrization $w = 2w'$ in all linear layers appears as a great improvement. BN Eval curve shows performance on the complete training set when BN is switched to the 'evaluation' mode (batch statistics are not used).}
\end{figure}
In our first attempt of comparing normalized and unnormalized training, we compare different methods with equal learning rate, initialized as recommended by the specific method. An example of such a comparison is shown in~\cref{fig:naive}. It turns out to be rather uninformative and even misleading.
First, we are comparing training schemes starting at different points. Clearly, a better starting point gives an advantage~\cite{Mishkin-16-Init,Salimans2016WeightNA}. Thus, the effects of initialization and normalization are entangled.
Second, comparing at equal learning rates is incorrect because the methods use different parametrization and thus efficiently rescale the descent steps differently, as explained below. Third, with limited training sets, NNs easily over-fit and the training cross-entropy loss can be made arbitrary close to zero while having poor generalization. To remedy this we will consider noise-augmented training sets.

%
%The original work~\cite{IoffeS15} compared BN to the standard unnormalized training in a way, which we find not informative enough.
%First, BN is initialized without preserving equivalence, which means that the compared methods start at different points. While it may be practical, the effects of initialization and normalization during the training are not decoupled. Second, learning rate of BN was manually adjusted.
%The previous work with respect to this issue 

%One could think that comparing at equal learning rate is more consistent. This is however not true because of the difference in reparametrizations. 
To see the issue with the learning rate, consider what happens with the step size when a reparametrization is used.
Let for example, $\theta' = A \theta$ (a linear change of variables). The steepest descent in $\theta$ for an objective $f(\theta)$ has the form $\theta^{t+1} := \theta^{t} - \alpha_t \nabla f(\theta)$. The steepest descent in $\theta'$ for the minimization of $f(\theta) = f(A^{-1} \theta')$ can be equivalently written as
\begin{align}
\theta^{t+1} = \theta^{t} - \alpha_t (A A\T)^{-1} \nabla f(\theta^t),
\end{align}
\ie, as a preconditioned gradient descent (see \eg,~\cite[\protect\parSym 8.7]{Luenberger:2015:LNP}). If we keep the learning rate the same, a method that minimizes $f(2\theta')$ results in a 4 times larger step size. Compared as in~\cref{fig:naive}, it can be declared a ``faster normalization scheme''. The issues applies to normalization schemes because a change of variables is involved and there is also a possibility to, \eg, overestimate the variance.
\par
The previous work~\cite{IoffeS15} compared BN to the standard unnormalized training initialized differently and manually adjusting learning rate and other parameters.
These issues are better addressed in~\cite{Salimans2016WeightNA}, where BN, WN and an unnormalized network are all initialized with BN and a set of learning rates is tried for each method ($0.0003, 0.001, 0.003, 0.01$).
A more recent work~\cite{Gitman-17} used a fixed initial learning rate and tuned the decay rule of learning rate of some methods manually.
Furthermore,~\cite{IoffeS15,Gitman-17} apply weight decay ($l_2$ regularization of weights) to BN (footnote 3 in~\cite{Gitman-17}) while being aware of the scale invariance discussed in~\cite{IoffeS15}. Adding to the objective $\lambda \|w\|_2^2$, however small the coefficient is, makes the problem ill-conditioned. The weights can be taken arbitrary close to zero without affecting the cross-entropy while decreasing $\|w\|_2^2$, with a singularity of the objective at zero. It follows that there is no minimum and such regularization achieves nothing but destabilizing the learning\footnote{
The gradient descent may however behave well in some cases. As explained in~\cite{Salimans2016WeightNA}, due to scale-invariance the gradients of the main loss are orthogonal to $w$ and the steps monotonously increase the norm $\|w\|$. The regularization can be balancing this effect. However, a projection on the constrain $||w|| = 1$ would be much simpler.
%Because of the invariance to $\|w\|$ the gradient of the main objective is orthogonal to $w$. The steps orthogonal to the current vector $w$ monotonously increase its norm $\|w\|$, as explained in~\cite{Salimans2016WeightNA}. 
%The two effects may be compensating each other. However, one can simply project on the constraint after each iteration instead of balancing with weight regularization. 
Weight regularization of parameters $s, b$ is also doubtful. Non-linearities introduce an implicit scaling and bias depending on previous layers. Making $\|b\|$ smaller does not necessarily make the total effective bias smaller, but instead can make it larger.
}.
\section{Experiments}
Similarly to~\cite{Salimans2016WeightNA} we compare different methods starting from the same point defined as follows:
\begin{itemize}
\item init=None: Initialize weights randomly, normalizations are introduced with preserving the equivalence.
\item init=BN: BN is introduced with $s$ random and $b=0$ (pytorch default), projecting out scale and bias of the original netowrk. mean and variance of a single batch are used to convert the model to unnormalized form for starting of other methods.
\item init=AP2: The proposed normalization is introduced with $s=1$, $b=0$ and then converted to the unnormalized form for starting of other methods.
\end{itemize}
Normalization is introduced after every linear (conv or fully connected) layer.

To select the learning rate we propose to numerically optimize it for the best training objective in the horizon of 5 epochs. We do not try a set of values but instead apply a zero order optimization method for a function of a single variable (the learning rate) as detailed below. While it does not fully resolve the issue of a fair comparison, it is a method that can be used in practice and it does make the comparison more independent of trivial reparametrizations.
%Since the objective is stochastic, what we optimize is its running mean.

\paragraph{Implementation Details}
Our implementation in pytorch will be made publicly available.

To find the best learning rate \textit{lr} we use the bounded Brent's method\footnote{Available in {\tt scipy} package} to optimize the $\log_{10} {\textit lr}$ with bounds $[-6,\,-2]$ and a limit of 10 iterations. The objective is the running mean estimate of the stochastic loss in five epochs. The running mean uses exponential weights vanishing to $0.1$ in 1 epoch.

\begin{figure}[!t]
\includegraphics[width=\linewidth]{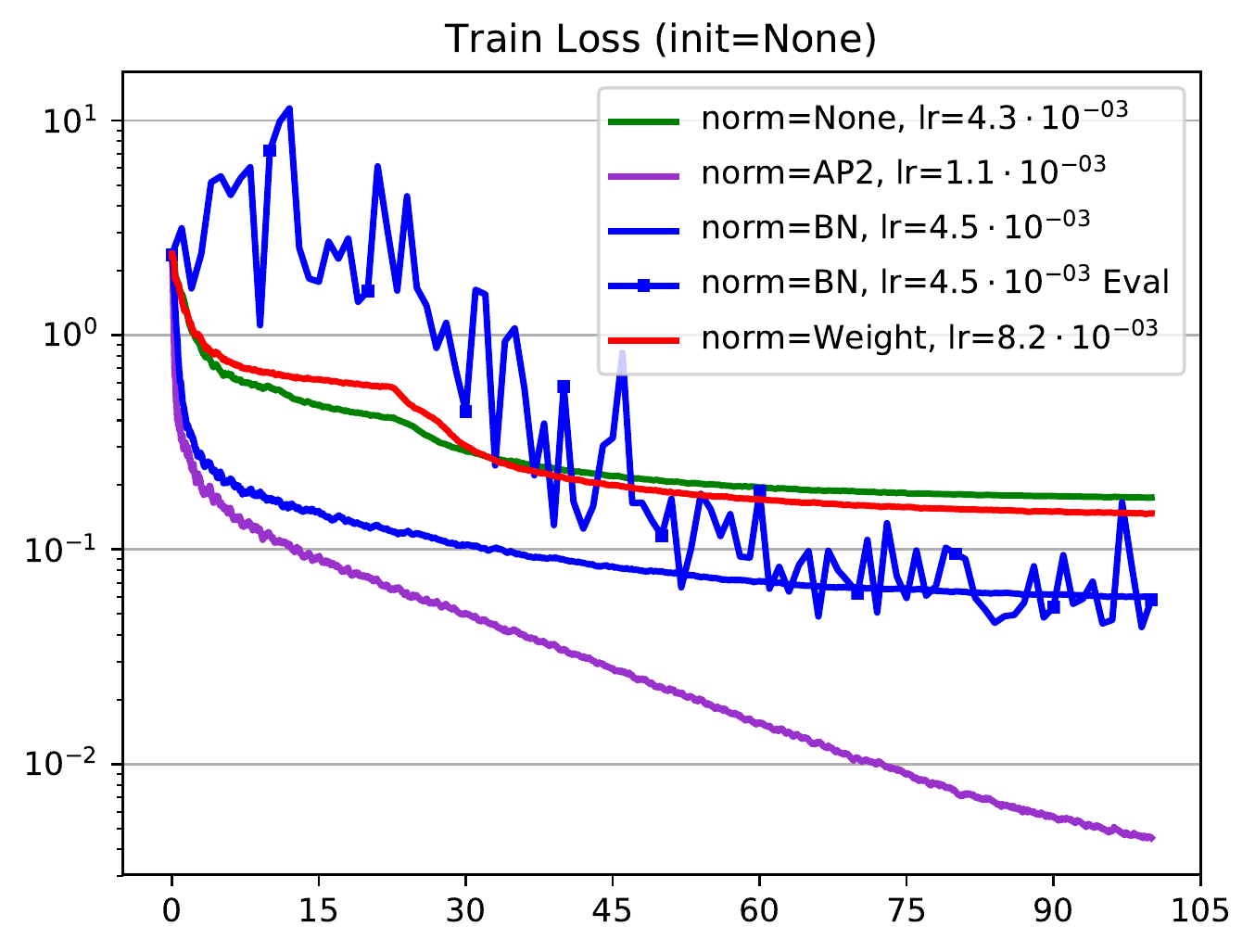}\\[5pt]
\includegraphics[width=\linewidth]{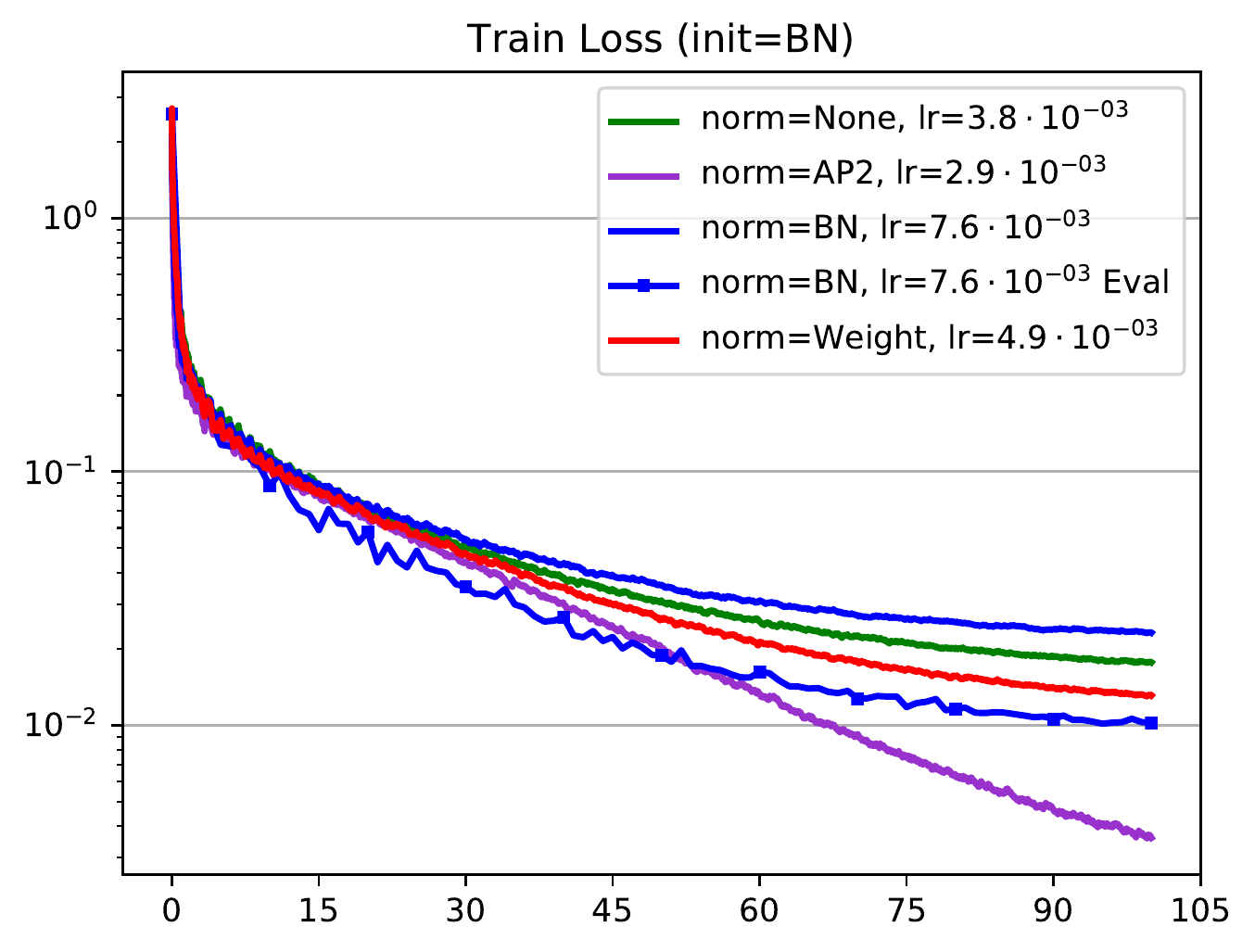}\\[5pt]
\includegraphics[width=\linewidth]{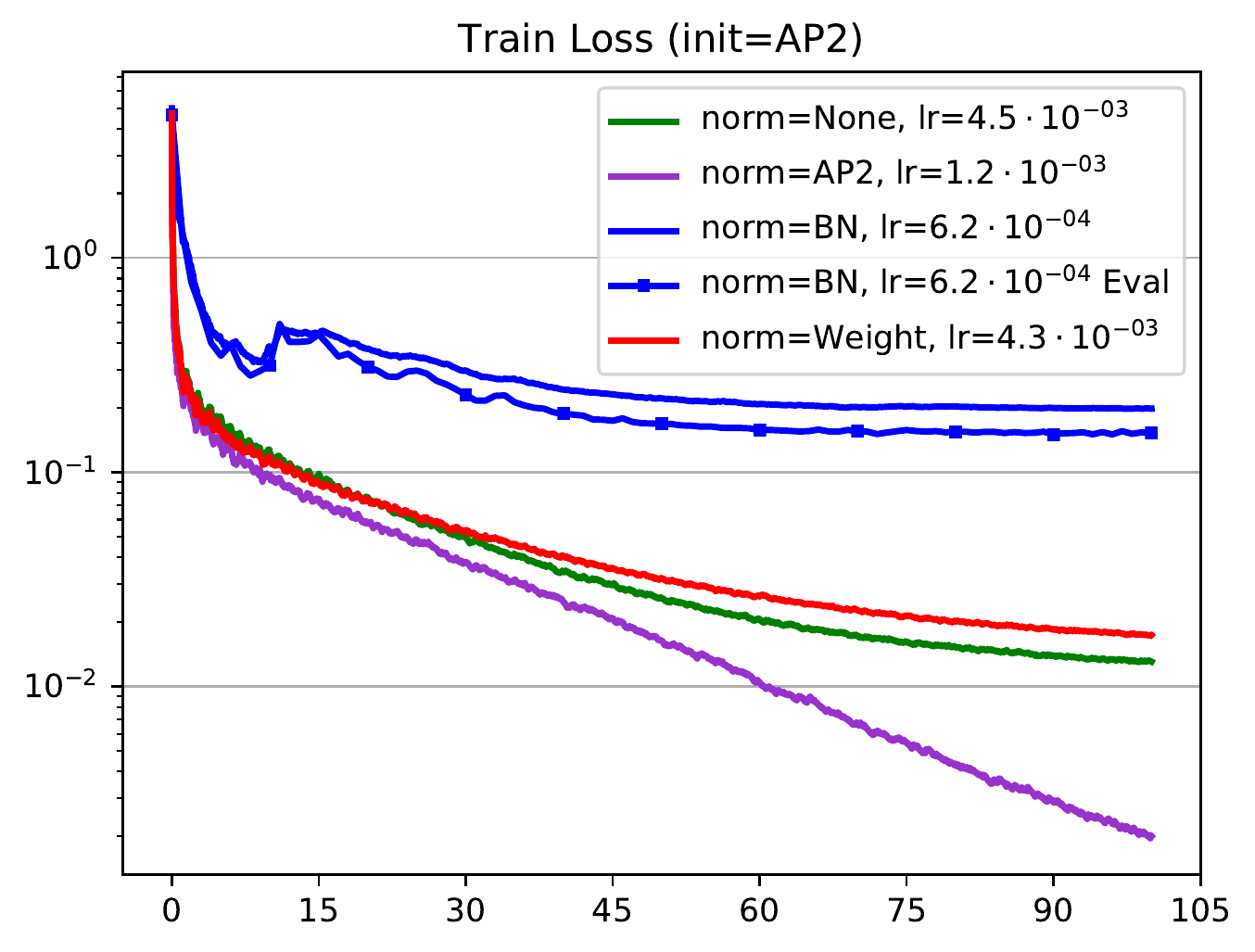}
\caption{\label{fig:MNIST}
MNIST (no noise augmentation). Plots are grouped by the initialization method and show a running mean estimate of the stochastic batch-wise training objective \vs epochs. For all normalizations but BN this matches well with the training loss on the whole data set. For BN, its stochastic estimate deviates from its evaluation performance on the training set (shown as a Eval). In this task it is possible to bring the loss arbitrary close to zero, which AP2 normalization successfully does for all 3 inits -- a clear case of overfitting.
}
\end{figure}
\begin{figure}[!t]
\includegraphics[width=\linewidth]{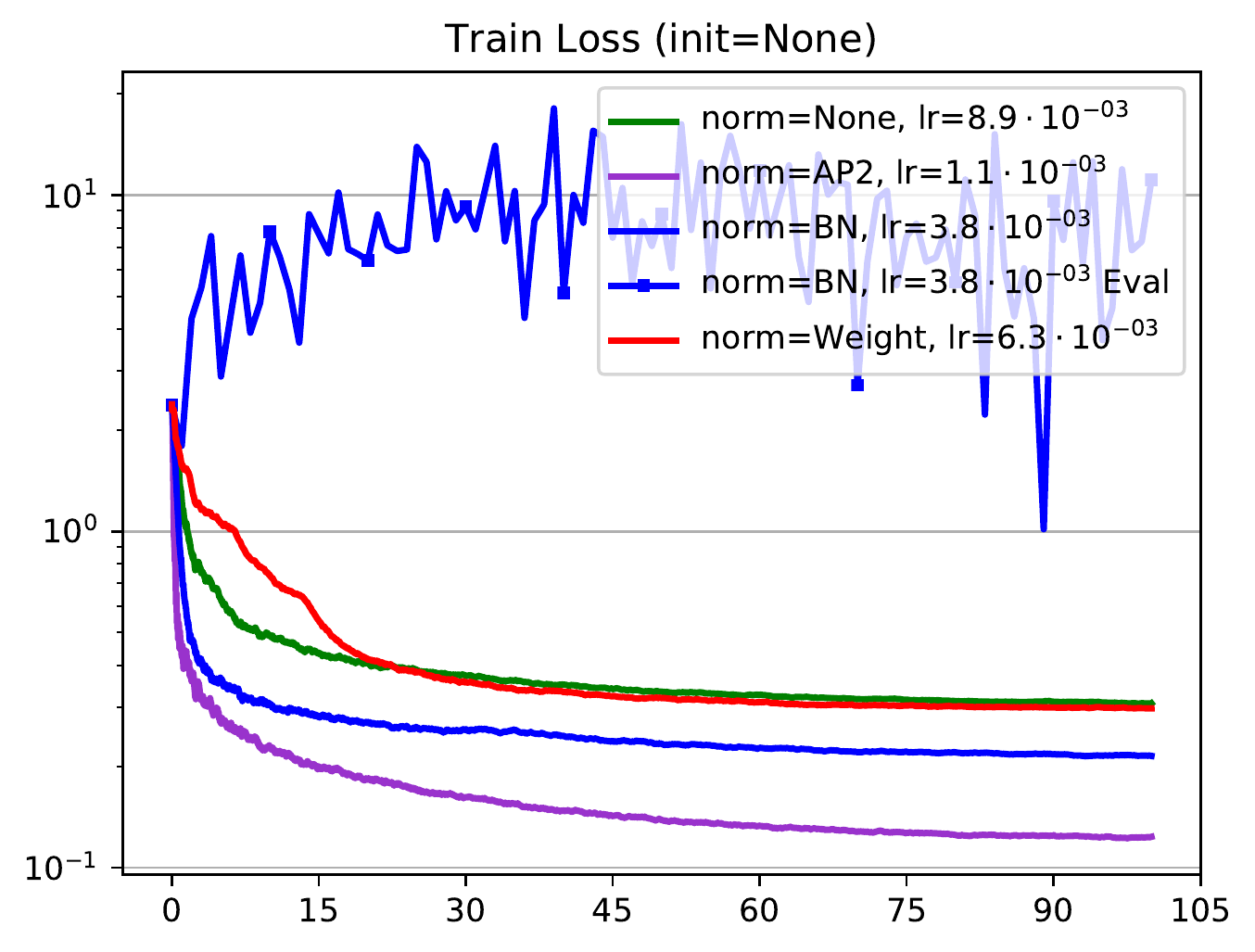}\\[5pt]
\includegraphics[width=\linewidth]{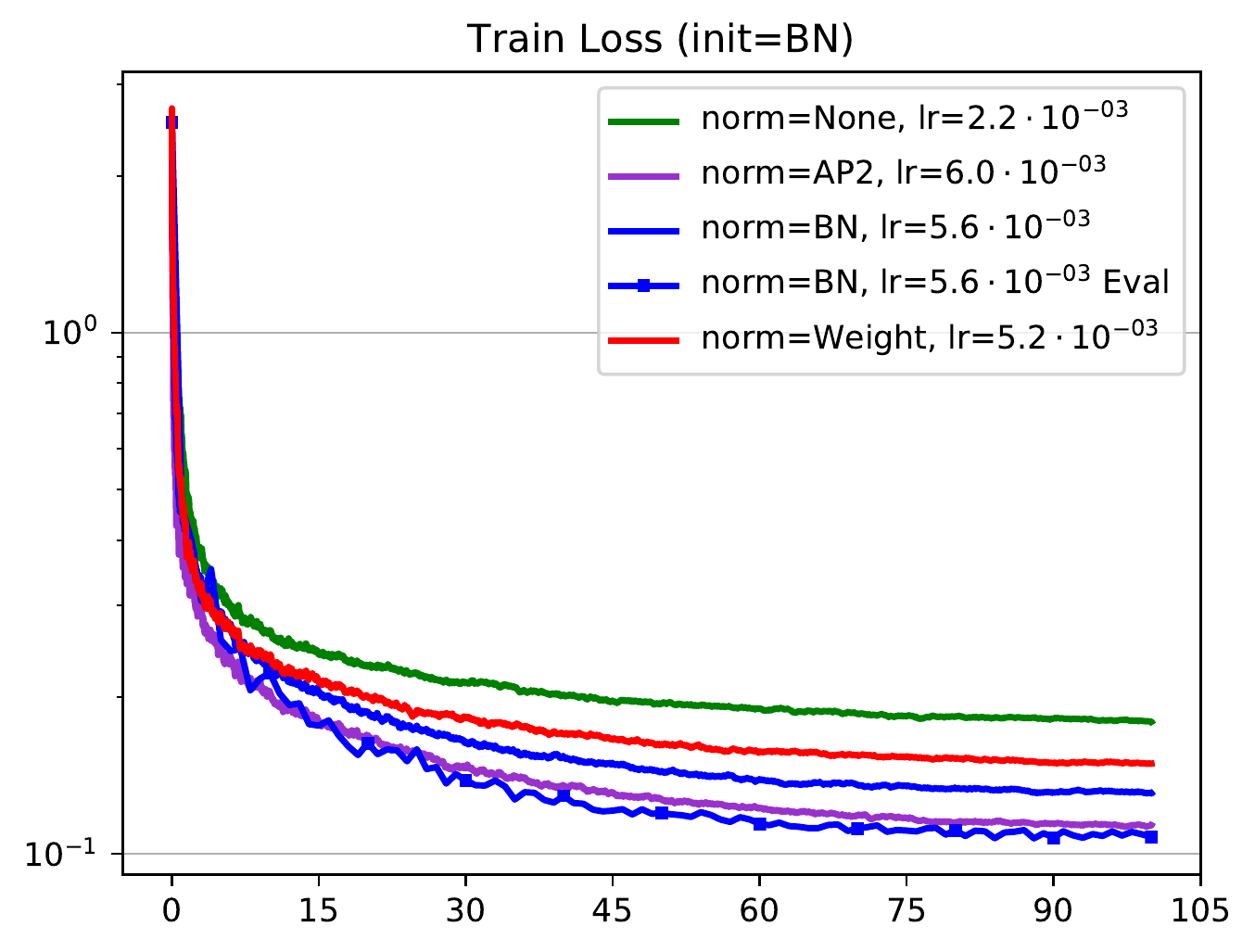}\\[5pt]
\includegraphics[width=\linewidth]{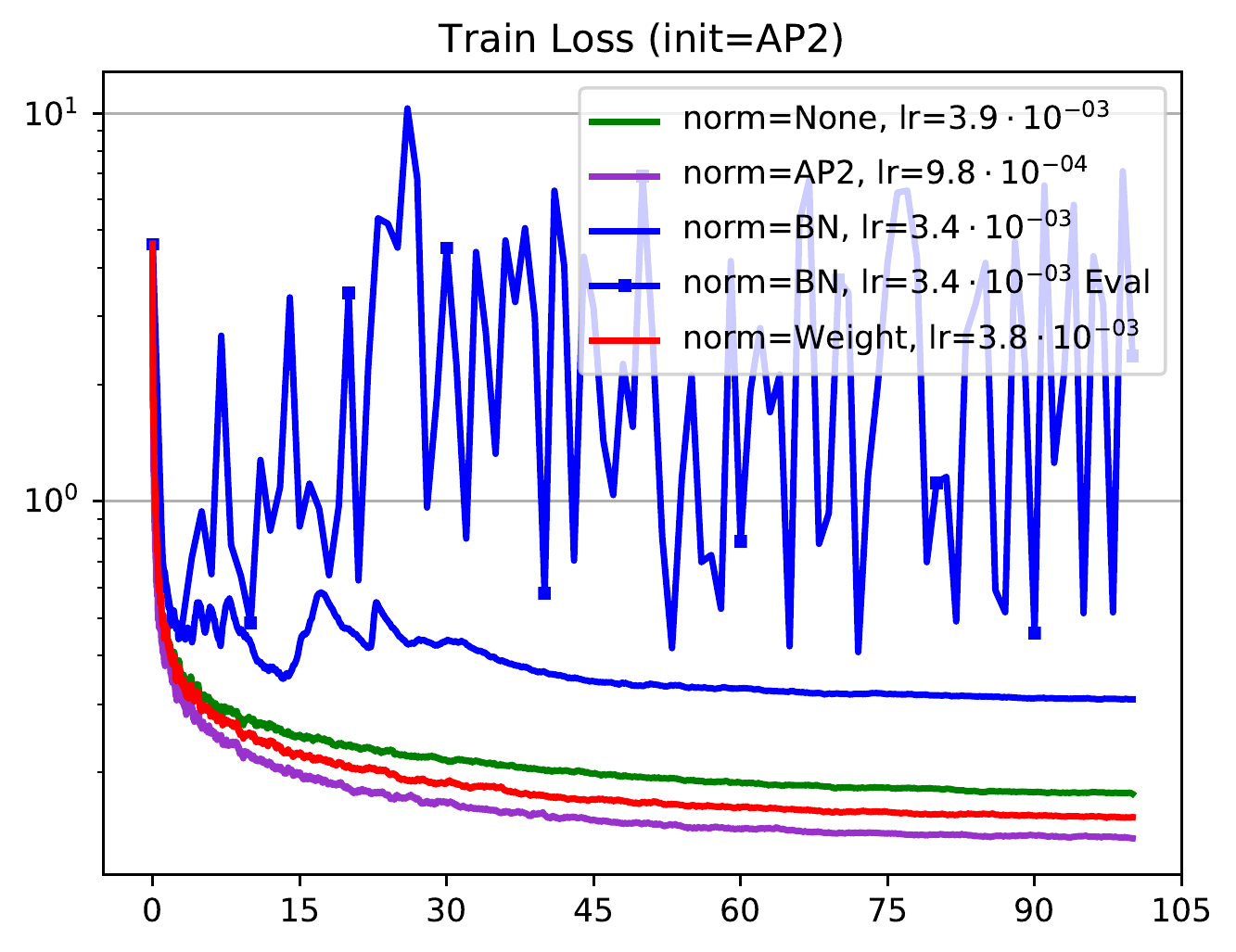}
\caption{\label{fig:MNISTn}
Noisy MNIST: Training loss with different initializations.
}
\end{figure}

\paragraph{Parameters}
We used batch size 128, Adam optimizer with learning rate ${\textit lr} \cdot 0.96^k$, where $k$ is the epoch number (this gives a factor 10 reduction in 50 
epochs). The initial learning rate $lr$ is the parameter optimized as discussed above.
BN parameters are the pytorch default ones: {\em eps} $=10^{-5}$, {\em momentum}\footnote{In pytorch this is the weight of the new data point, not of the previous estimate.} $=0.1$.

\paragraph{Datasets}
We used MNIST\footnote{\url{http://yann.lecun.com/exdb/mnist/}} and CIFAR10\footnote{\url{https://www.cs.toronto.edu/~kriz/cifar.html}} datasets. %Both dataset provide a split into training and test sets. From the training set we split 10 percent (at random) as a validation set.
\begin{figure*}[!t]
\includegraphics[width=0.33\linewidth]{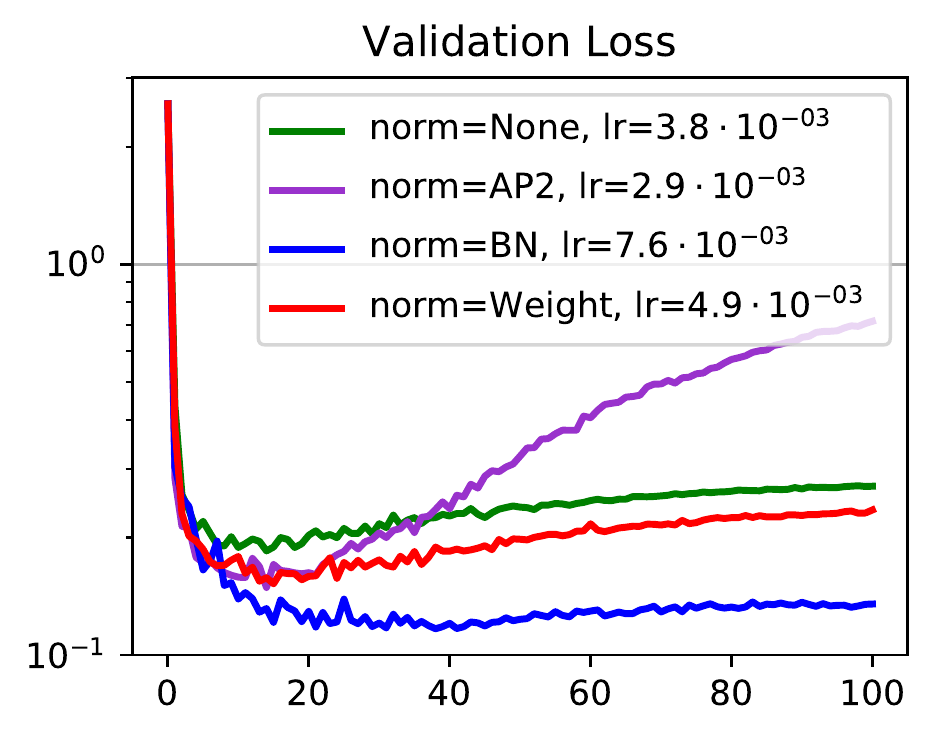}\ %
\includegraphics[width=0.33\linewidth]{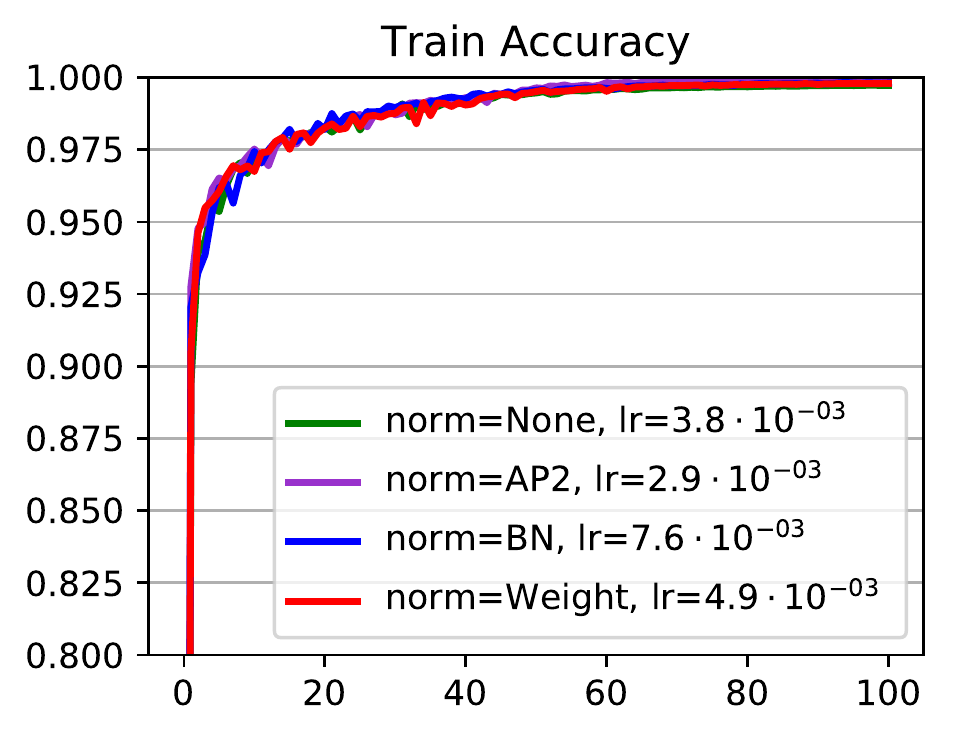}\ %
\includegraphics[width=0.33\linewidth]{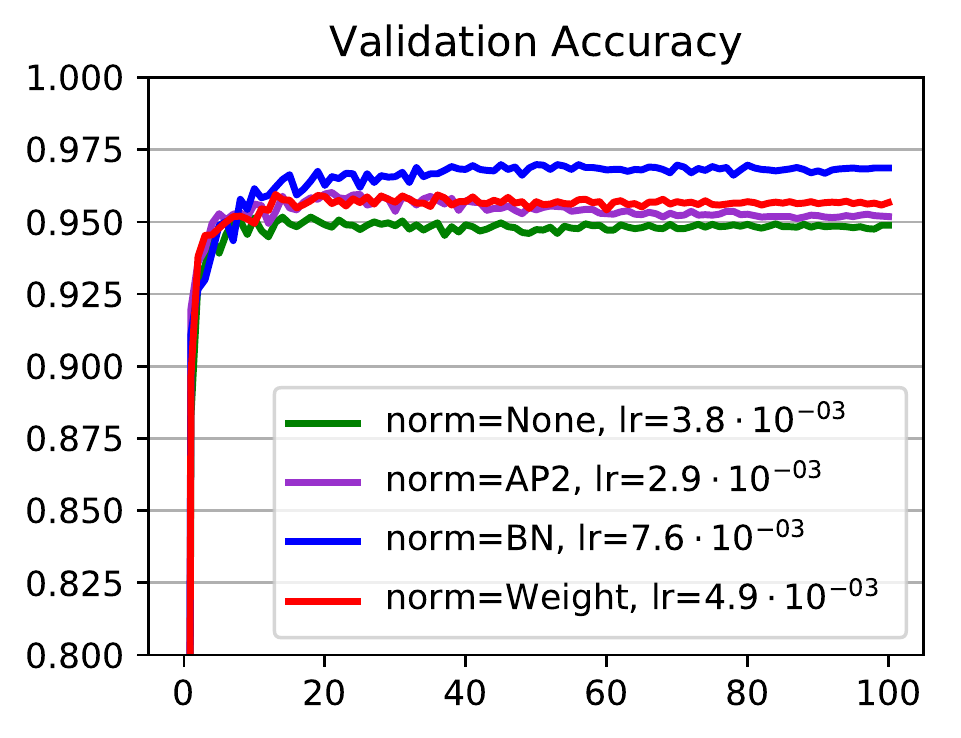}%
\caption{\label{fig:MNIST-v}
MNIST: Validation loss and accuracies, BN-initialized training. BN shows less overfitting and better validation accuracy.
These results are consistent with~\cite[figure 3]{Gitman-17}.}
\end{figure*}
\begin{figure*}[!t]
\includegraphics[width=0.33\linewidth]{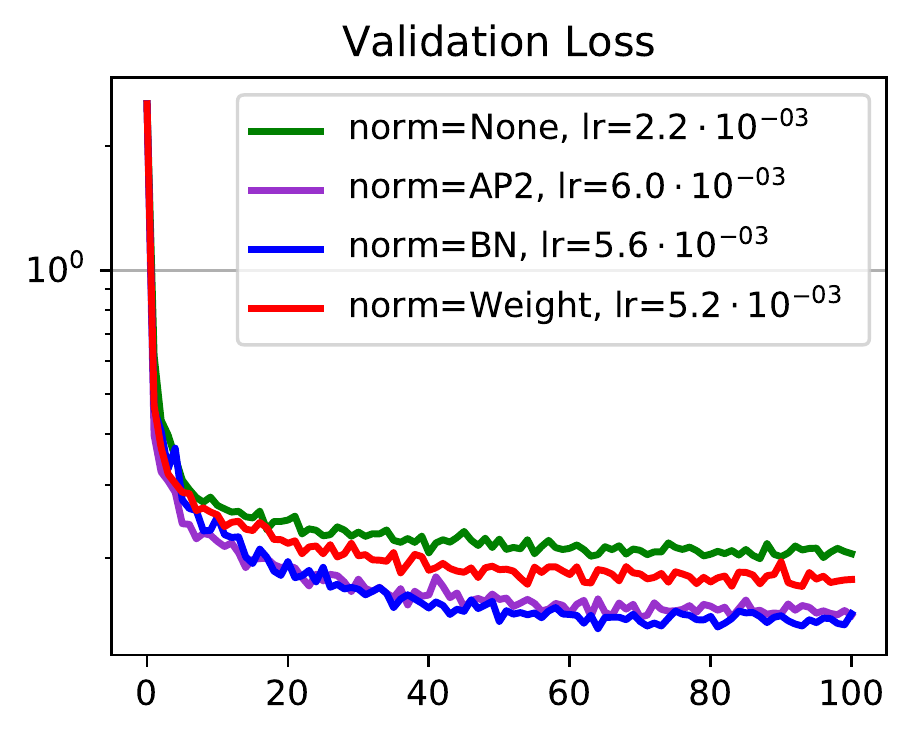}\ %
\includegraphics[width=0.33\linewidth]{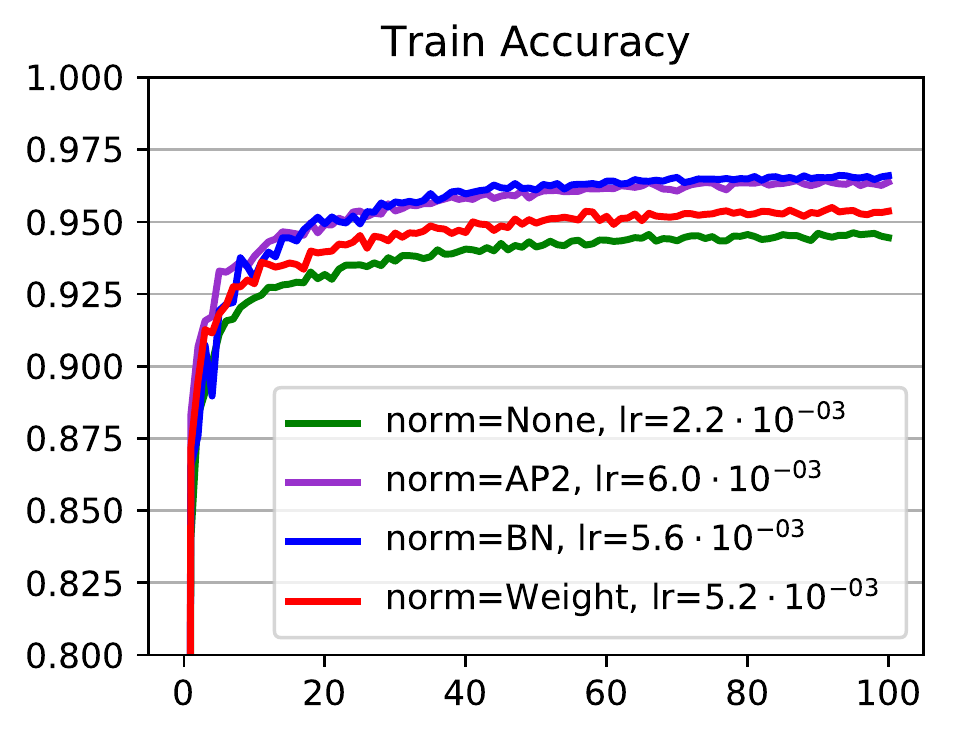}\ %
\includegraphics[width=0.33\linewidth]{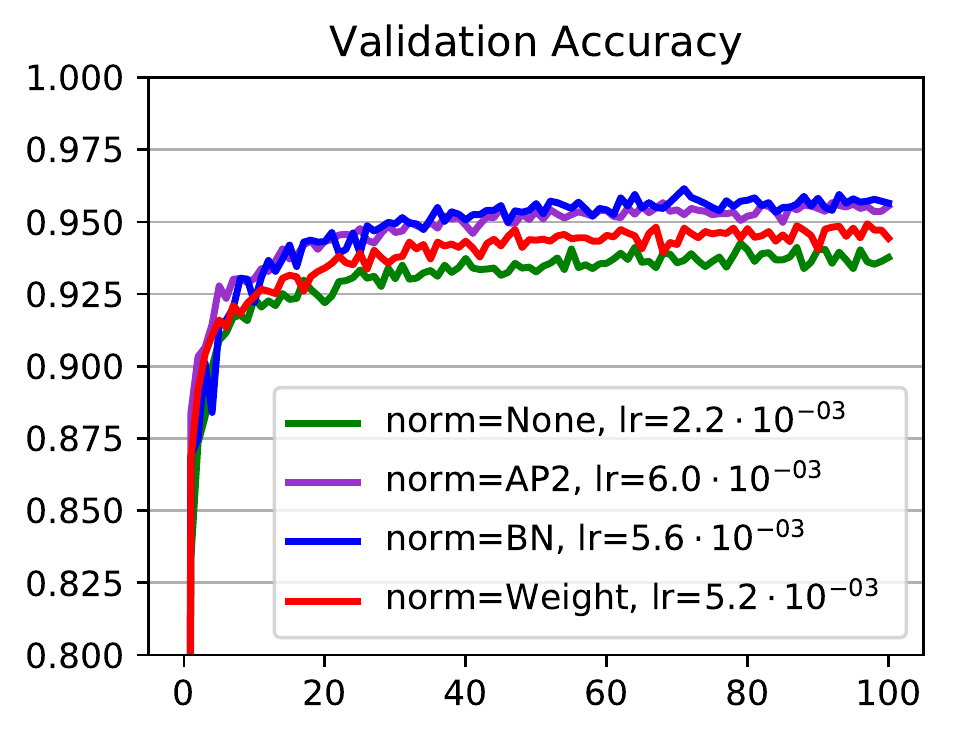}%
\caption{\label{fig:MNISTn-v}
Noisy MNIST: Validation loss and accuracies, BN-initialized training. Here overfitting does not occur and AP2 is on par with BN.
}
\end{figure*}
\begin{figure*}[!t]
\includegraphics[width=0.33\linewidth]{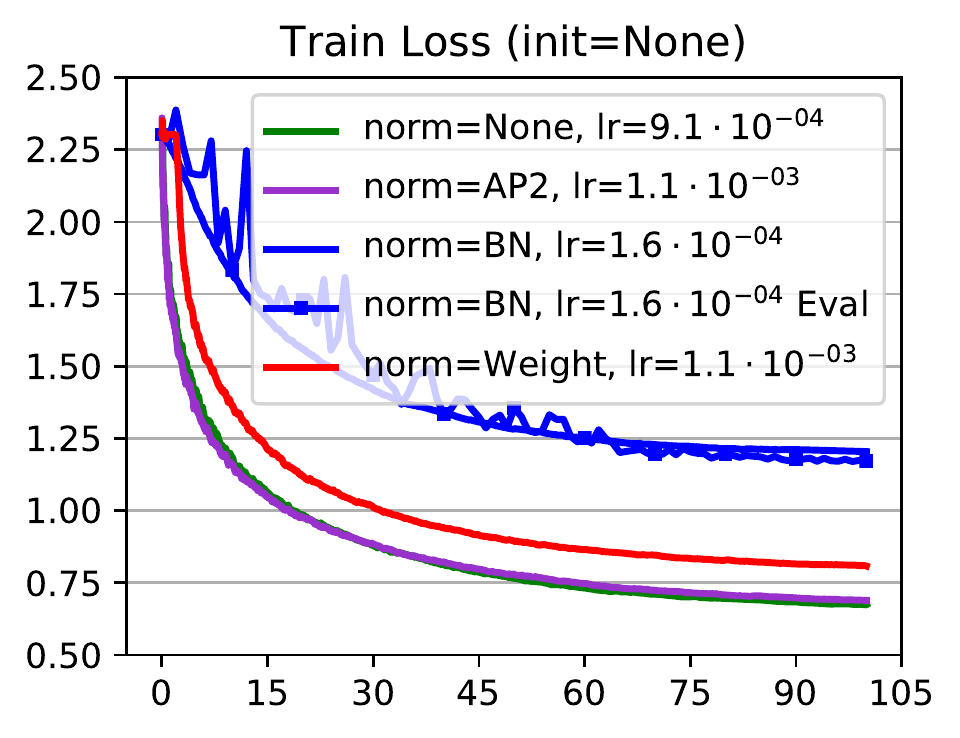}\ %
\includegraphics[width=0.33\linewidth]{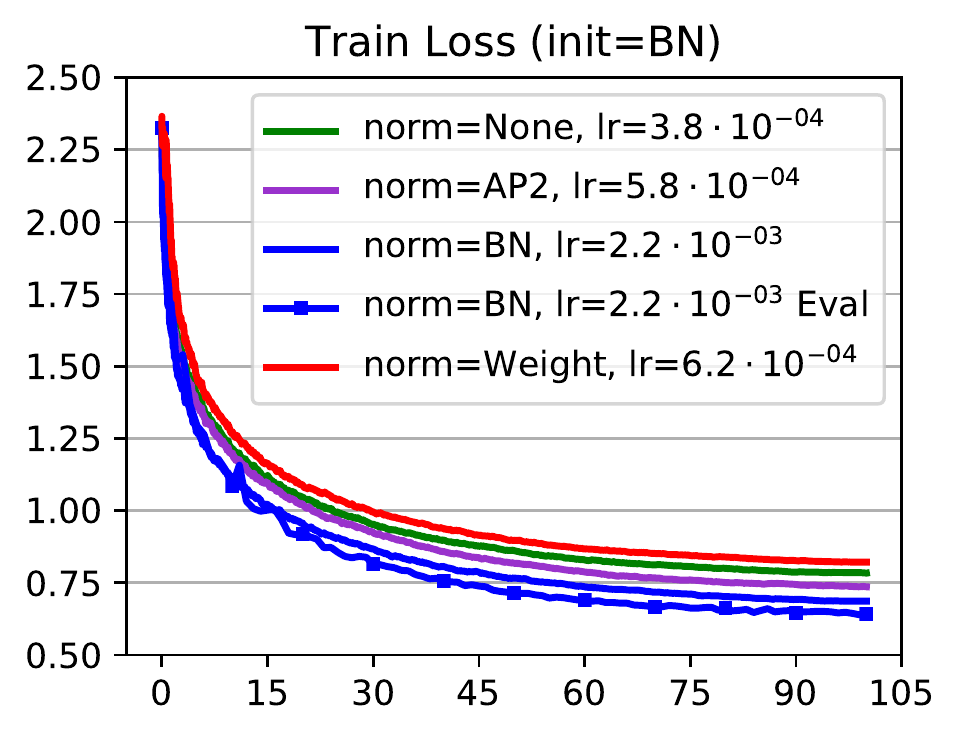}\ %
\includegraphics[width=0.33\linewidth]{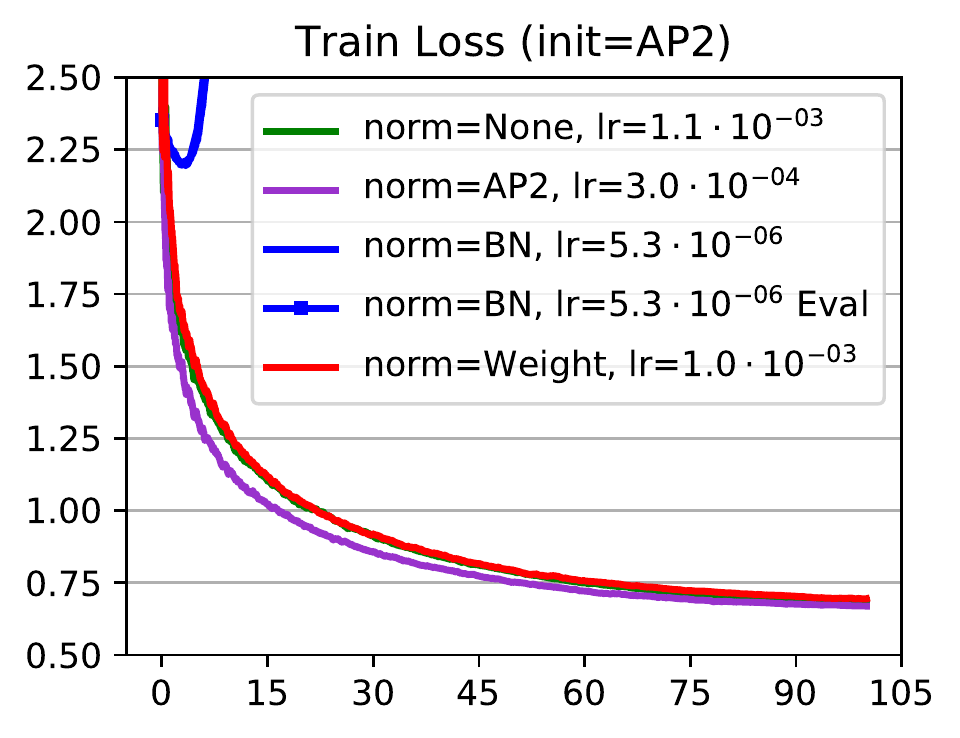}%	
\caption{\label{fig:CIFARn}
Noisy CIFAR-10: Training loss with different initializations. When the learning rate is optimized as proposed in the paper, the differences between different init and normalization schemes are not so pronounced and we cannot make conclusions about the speed of learning.
}
\end{figure*}
\begin{figure*}[!t]
\includegraphics[width=0.33\linewidth]{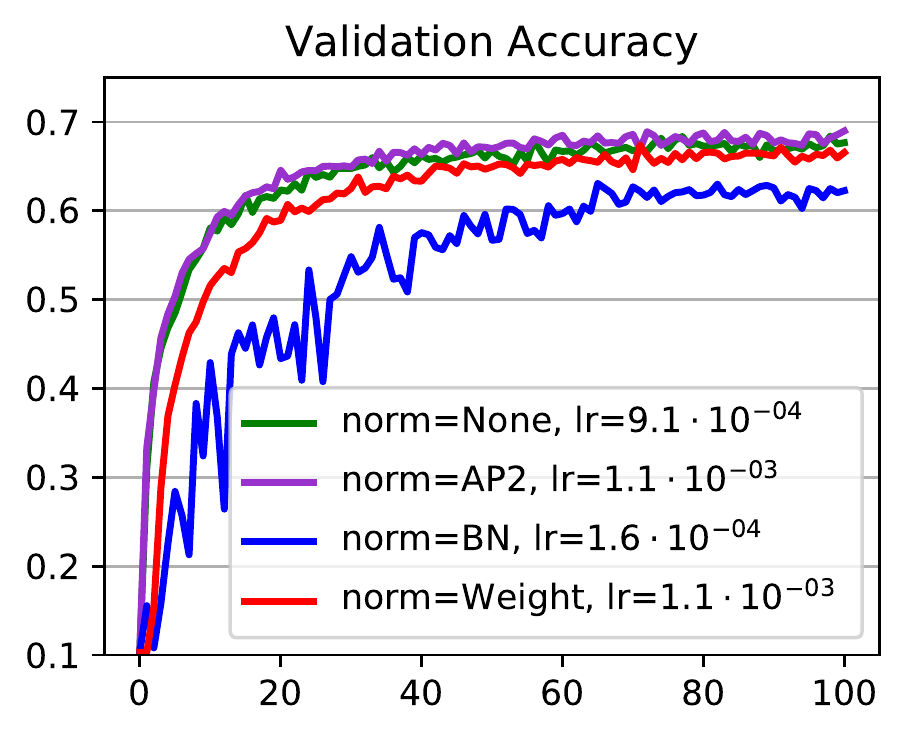}\ %
\includegraphics[width=0.33\linewidth]{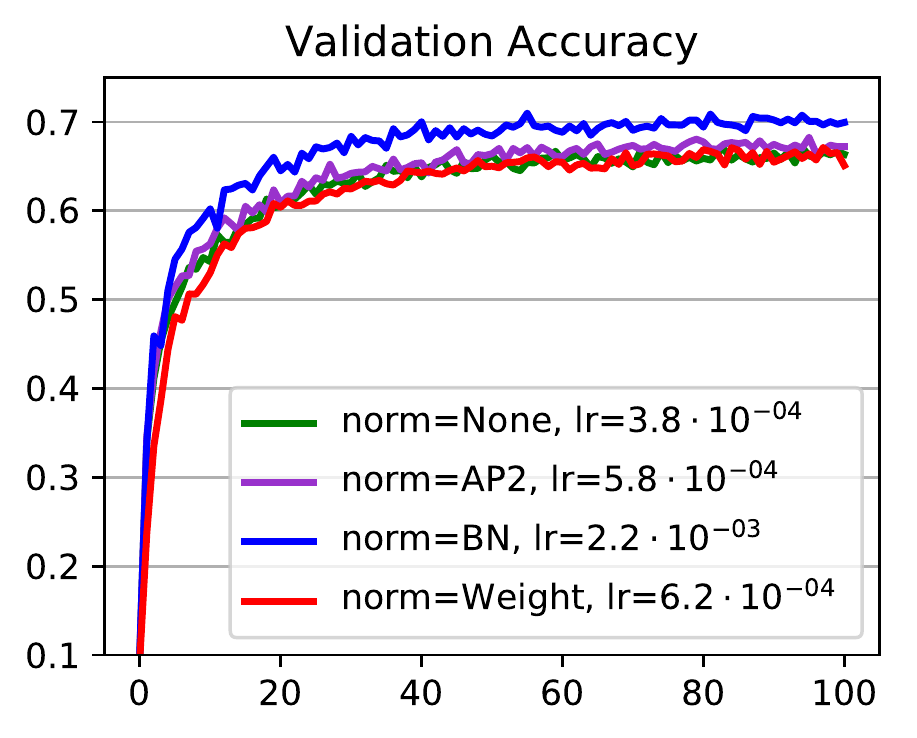}\ %
\includegraphics[width=0.33\linewidth]{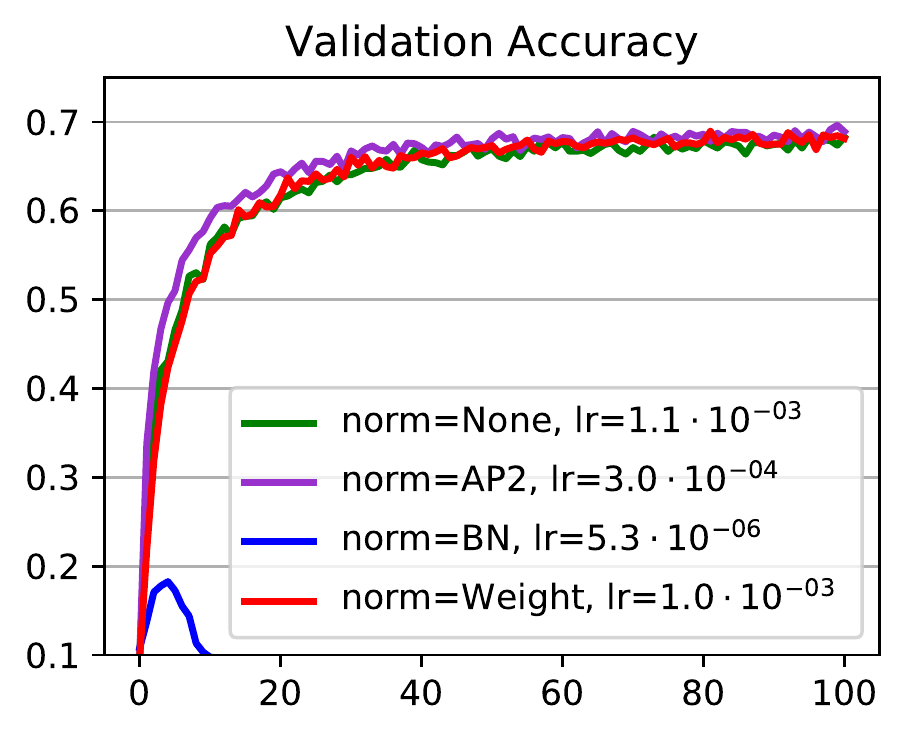}%
\caption{\label{fig:CIFARn-v}
Noisy CIFAR-10: Validation accuracies corresponding to \cref{fig:CIFARn}.
\skipmyskip
\skipmyskip
\skipmyskip
\skipmyskip
\skipmyskip
\skipmyskip
}
\end{figure*}
\paragraph{MNIST}
The initial dataset is augmented by random offsets in the range $[-2,2]$ with zero padding.
We tested a simple non-convolutional network with 6 hidden layers of logistic transform units, 20 in each (same as in~\cite{Shekhovtsov-17}).
Such network is already relatively hard to train with random initialization as seen in~\cref{fig:MNIST} (top), even when the best learning rate is chosen. BN starting from this initialization performs better but has problems with estimating the whole training set mean and variance. Its evaluation performance is very stochastic and does not match the training performance\footnote{Perhaps a smaller value of {\em momentum} is needed. This hyper-parameter however does not affect the training process.}.
With BN initialization, all methods succeed to train the network. 
It is seen that with optimized learning rates, their performance is much more similar than reported elsewhere. Here BN evaluation-mode  performance is better than its batch performance.
Finally, for AP2 initialization, standard network and weight normalization work well while BN oscillates despite that the automatically chosen learning rate is by an order of magnitude smaller than in the case of {\tt init=BN}.
It is interesting that with AP2 normalization, the learning objective is optimized well for all initializations. However, in this test scenario the model is overfitting, as can be seen from the validation loss and accuracy shown in~\cref{fig:MNIST-v}, and therefore a better performance on the training loss is of little practical utility.
\paragraph{Noisy-MNIST}
The dataset is additionally augmented with Gaussian noise with variance $0.1$.
The results with the same model and settings as MNIST are presented in \cref{fig:MNISTn,fig:MNISTn-v}.

\paragraph{Noisy-CIFAR-10}
We tested with a CNN network with ReLU and the following conv layers:
\begin{lstlisting}[basicstyle=\scriptsize\ttfamily]
ksize =[3,  3,  3,  3,   3,   3,   3,   1,   1 ]
stride=[1,  1,  2,  1,   1,   2,   1,   1,   1 ]
depth =[96, 96, 96, 192, 192, 192, 192, 192, 10]
\end{lstlisting}
listing kernel size, input stride, and output channels, resp. We used a leaky ReLU activation: $\max(0,x) + 0.03x$. The reason is that with pure ReLU some methods were finding a local minimum that detached the input (\eg, when one layer becomes fully saturated at 0 and the others fit just the prior class distribution), which was breaking our learning rate optimization. LReLU follows all layers but the last one, followed by log softmax. The initial dataset is augmented by random horizontal flips, offsets in the range $[-2,2]$ with zero padding and noise with variance $0.1$.

The results are shown in~\cref{fig:CIFARn}. The proposed normalization behaves well across all init points. BN achieves the overall best training objective, but the difference is not decisive, comparable to results of AP2 with other inits. The same holds for the validation accuracies in~\cref{fig:CIFARn-v}. BN has a significant generalization boost when using accumulated statistics (evaluation mode). This is an interesting phenomenon, perhaps related to boosting or ensemble techniques: the data-dependent (stochastic) normalization values are replaced with their running average counterparts (expectations).

%\cite{Gitman-17}

\section{Conclusion}
In this work we explored an application of variance propagation~\cite{Shekhovtsov-17} for computing normalization statistics.
The resulting method has similar advantages as weight normalization: computation efficiency, continuous differentiability, data-independent plug-in/out capability, applicability to recurrent nets, etc. The procedure approximates the needed statistics in a clearly understood manner and is general in the sense that it can take into account different constructive elements of neural networks and their dependencies.
We strove towards making experiments more objective by considering same initialization points and optimizing the learning rate. We observed that the remaining differences in the speed of different methods became much smaller than when compared naively, \eg, as in~\cref{fig:naive}.
It can be still observed that the proposed technique improves robustness to initialization points and achieves a lower training objective in many cases. It also gives a good initialization point to standard parametrization and weight normalization. It does not have the property of batch normalization to generalize (presumably due to stochasticity and averaging). The later however experienced a significant instability when applied to other initialization points in our tests.

We hypothesize that the proposed method may benefit from an orthogonal initialization because such initialization would make the uncorrelated inputs assumption exact. The role of the norm $\|w\|$, to which the normalized forms are invariant (but not their gradients!) is yet unclear.

\section*{Acknowledgment}
We thank Tomas Werner and Dmytro Mishkin for discussions and reviewers for their feedback. A.~Shekhovtsov was supported partially by Toyota Motor Europe HS and partially by Czech Science Foundation grant 18-25383S.  B.~Flach was supported by Czech Science Foundation under grant 16-05872S.
{\small
\bibliographystyle{apa}
\bibliography{../../bib/strings,../../bib/neuro-generative,../../bib/our}
}

\end{document}

%% file: style/packages.tex
\usepackage{etex}
\usepackage{url}
\usepackage[numbers]{natbib}

\usepackage{cvpr-abbriv}
\usepackage{epsfig}
\usepackage{graphicx}
\usepackage{amsmath}
\usepackage{amssymb}
\usepackage{array}
\usepackage{url}
\usepackage{graphicx}
\usepackage{amsmath,amssymb} % define this before the line numbering.
\makeatletter
\let\proof\@undefined                        % undefine \proof
\let\endproof\@undefined                  % undefine \endproof
\makeatother
\usepackage{amsthm}
\usepackage{amsxtra}
\usepackage{stmaryrd}
\usepackage{comment}
\usepackage[ruled,vlined,linesnumbered,procnumbered,longend]{algorithm2e}

\SetCommentSty{mycommfont}
\usepackage{setspace}
\usepackage{booktabs}
\usepackage{color, colortbl}
\usepackage{tikz}                 % create images using pgf/tikz
\usetikzlibrary{patterns,fit,external}
\usepackage{pgfplots}             % draw plots in tikz style
\usepackage{overpic}
\usepackage{tocloft}
\usepackage{bbm}
\usepackage{atbeginend}
\usepackage{units}
\usepackage{xifthen}
\usepackage{listings}

\usepackage[titletoc,title]{appendix}

\usepackage{times}
\usepackage{tikzscale}
\usepackage[textwidth=2cm,figwidth=\columnwidth,colorinlistoftodos]{todonotes}
\makeatletter
\renewcommand{\todo}[2][]{\tikzexternaldisable\@todo[#1]{#2}\tikzexternalenable}
\makeatother

\newcounter{mycomment} % Usage:  \mycomment[CR]{Schreib was gscheits}

\usepackage{booktabs}
\usepackage[
  position=bottom,
  font=small,
  labelfont={bf},
]{caption,subfig}
\usepackage{tabu} %% !!! this is better!
\usepackage[most]{tcolorbox}
\usepackage{pbox}
\newlength{\luw}
\newlength{\luh}

\usepackage{xifthen}

\usepackage{enumitem}

\usepackage[draft=false, pagebackref=false,breaklinks=true,colorlinks,bookmarks=false,pdftex]{hyperref}
\hypersetup{hidelinks,colorlinks=false,linkbordercolor={1 1 1}}
\input{style/tex_defs.tex}
\input{style/myspace.tex}
\usepackage{thmtools, thm-restate}
\usepackage[capitalise,nameinlink]{cleveref}
\crefformat{equation}{(#2#1#3)}
\crefname{section}{\parSym}{\parSym\parSym}
\Crefname{section}{\parSym}{\parSym\parSym}
\crefname{appendix}{\parSym}{\parSym\parSym}
\crefname{observation}{Observation}{Observations}
\usepackage[bottom]{footmisc}
\usepackage{chngcntr}

%% file: style/tex_defs.tex
\DeclareMathAlphabet{\mathcalligra}{T1}{calligra}{m}{n}
\DeclareMathAlphabet{\mathantt}{OT1}{antt}{li}{it}
\DeclareMathAlphabet{\mathpzc}{OT1}{pzc}{m}{it}

\DeclareMathOperator{\Var}{Var}
\DeclareMathOperator{\Cov}{Cov}

\newcommand{\<}{\langle}
\renewcommand{\>}{\rangle}

\newcommand{\R}{\mathcal{R}}

\newcommand{\E}{\mathbb{E}}

\newcommand{\N}{\mathcal{N}}

\def\T{^{\mathsf T}}

\newcounter{myRomanCounter}

\newcommand{\gray}{\color[rgb]{0.5,0.5,0.5}}
\newcommand{\red}{\color[rgb]{1,0,0}}

\renewcommand*{\paragraph}[1]{\par{\normalsize\bf #1}\,\xspace}

\def\epsilon{\varepsilon}

\makeatletter
\let\parSym\S
\def\S{\mathcal{S}}
\makeatother


\newcommand{\revisit}[1][]{%
\ifthenelse{\equal{#1}{}}{% no argument
\ensuremath{\red \triangle}\xspace}{%
{\ensuremath{\red \rhd}\xspace}%
{\gray #1}%
{\ensuremath{\red \lhd}\xspace}%
}%
}

\def\anchor [#1]#2{%
\phantomsection{}#1\label{#2}%
\def\arga{#2}%
\global\expandafter\def\csname#2\endcsname{%
\hyperref[#2]{#1}\xspace%
}%
}%

\def\codefunction [#1]#2{%
\phantomsection{}\label{#2}{\ttfamily #1\xspace}%
\def\arga{#2}%
\global\expandafter\def\csname#2\endcsname{%
\hyperref[#2]{\ttfamily #1}\xspace%
}%
}

\usepackage{array}
\newcolumntype{L}[1]{>{\raggedright\let\newline\\\arraybackslash\hspace{0pt}}m{#1}}
\newcolumntype{C}[1]{>{\centering\let\newline\\\arraybackslash\hspace{0pt}}m{#1}}
\newcolumntype{R}[1]{>{\raggedleft\let\newline\\\arraybackslash\hspace{0pt}}m{#1}}

\SetKwFor{forever}{while true}{}{end while}

\def\epsilon{\varepsilon}

\def\={\,{=}\,}

%% file: style/myspace.tex
\newlength{\myskip}
\def\skipmyskip{\vspace{0.5\myskip}}

\SetAlgoSkip{skipalgomyskip}
\SetAlgoInsideSkip{}
\makeatletter
\let\corollary\@undefined
\let\endcorollary\@undefined
\let\definition\@undefined
\let\enddefinition\@undefined
\let\proof\@undefined
\let\endproof\@undefined
\let\theorem\@undefined
\let\c@theorem\@undefined
\let\endtheorem\@undefined
\let\lemma\@undefined
\let\endlemma\@undefined
\let\example\@undefined
\let\c@example\@undefined
\let\endexample\@undefined
\let\remark\@undefined
\let\endremark\@undefined
\let\proposition\@undefined
\let\endproposition\@undefined
\let\property\@undefined
\let\endproperty\@undefined
\makeatother

\usepackage{thmtools}

\newtheoremstyle{tightItalic}% name
  {0.5\myskip}%      Space above
  {0\myskip}%      Space below
  {}%         Body font
  {}%         Indent amount (empty = no indent, \parindent = para indent)
  {\itshape}% Thm head font
  {.}%        Punctuation after thm head
  { }%     Space after thm head: " " = normal interword space;
  {}%         Thm head spec (can be left empty, meaning `normal')

\newtheoremstyle{tightBf}% name
  {0.5\myskip}%      Space above
  {0.5\myskip}%      Space below, 0 since amsart starts a paragraph
  {}%         Body font
  {}%         Indent amount (empty = no indent, \parindent = para indent)
  {\bf}% Thm head font
  {.}%        Punctuation after thm head
  {.5em}%     Space after thm head: " " = normal interword space;
  {}%         Thm head spec (can be left empty, meaning `normal')

\theoremstyle{definition}
\theoremstyle{tightBf}

\declaretheorem[style=tightBf,name=Theorem]{theorem}
\declaretheorem[style=tightBf,name=Lemma]{lemma}
\declaretheorem[style=tightBf,name=Definition]{definition}
\declaretheorem[style=tightBf,name=Corollary]{corollary}
\declaretheorem[style=tightBf,name=Proposition]{proposition}

\declaretheorem[style=tightBf,name=Example]{example}

\declaretheorem[style=tightItalic,name=Remark]{remark}

\theoremstyle{tightItalic}
\newtheorem*{proof}{Proof}
\BeforeEnd{proof}{\qed}


\setlist{leftmargin=0cm,itemindent=2ex,labelwidth=\itemindent,labelsep=0cm,align=left,parsep=0pt,itemsep=3pt,topsep=3pt}

%% file: main.bbl
\begin{thebibliography}{}

\bibitem[\protect\astroncite{Arpit et~al.}{2016}]{ArpitZKG16}
Arpit, D., Zhou, Y., Kota, B.~U., and Govindaraju, V. (2016).
\newblock Normalization propagation: {A} parametric technique for removing
  internal covariate shift in deep networks.
\newblock In Balcan, M. and Weinberger, K.~Q., editors, {\em ICML}, volume~48
  of {\em {JMLR} Workshop and Conference Proceedings}, pages 1168--1176.
  JMLR.org.

\bibitem[\protect\astroncite{Gitman and Ginsburg}{2017}]{Gitman-17}
Gitman, I. and Ginsburg, B. (2017).
\newblock Comparison of batch normalization and weight normalization algorithms
  for the large-scale image classification.
\newblock {\em CoRR}, abs/1709.08145.

\bibitem[\protect\astroncite{Hoffer et~al.}{2017}]{HofferHS17}
Hoffer, E., Hubara, I., and Soudry, D. (2017).
\newblock Train longer, generalize better: closing the generalization gap in
  large batch training of neural networks.
\newblock In Guyon, I., von Luxburg, U., Bengio, S., Wallach, H.~M., Fergus,
  R., Vishwanathan, S. V.~N., and Garnett, R., editors, {\em NIPS}, pages
  1729--1739.

\bibitem[\protect\astroncite{Ioffe}{2017}]{Ioffe17}
Ioffe, S. (2017).
\newblock Batch renormalization: Towards reducing minibatch dependence in
  batch-normalized models.
\newblock {\em CoRR}, abs/1702.03275.

\bibitem[\protect\astroncite{Ioffe and Szegedy}{2015}]{IoffeS15}
Ioffe, S. and Szegedy, C. (2015).
\newblock Batch normalization: Accelerating deep network training by reducing
  internal covariate shift.
\newblock In {\em ICML}, volume~37, pages 448--456.

\bibitem[\protect\astroncite{Klambauer et~al.}{2017}]{Klambauer-SELU}
Klambauer, G., Unterthiner, T., Mayr, A., and Hochreiter, S. (2017).
\newblock Self-normalizing neural networks.
\newblock {\em CoRR}, abs/1706.02515.

\bibitem[\protect\astroncite{{Lei Ba} et~al.}{2016}]{Ba-2016-Layer-Norm}
{Lei Ba}, J., {Kiros}, J.~R., and {Hinton}, G.~E. (2016).
\newblock {Layer Normalization}.
\newblock {\em ArXiv e-prints}.

\bibitem[\protect\astroncite{Liao et~al.}{2016}]{LiaoKP16}
Liao, Q., Kawaguchi, K., and Poggio, T.~A. (2016).
\newblock Streaming normalization: Towards simpler and more
  biologically-plausible normalizations for online and recurrent learning.
\newblock {\em CoRR}, abs/1610.06160.

\bibitem[\protect\astroncite{Luenberger and Ye}{2015}]{Luenberger:2015:LNP}
Luenberger, D.~G. and Ye, Y. (2015).
\newblock {\em Linear and Nonlinear Programming}.
\newblock Springer Publishing Company, Incorporated.

\bibitem[\protect\astroncite{Mishkin and Matas}{2016}]{Mishkin-16-Init}
Mishkin, D. and Matas, J. (2016).
\newblock {All you need is a good init}.
\newblock In {\em ICLR}.

\bibitem[\protect\astroncite{Ren et~al.}{2017}]{Ren-2016a}
Ren, M., Liao, R., Urtasun, R., Sinz, F.~H., and Zemel, R.~S. (2017).
\newblock Normalizing the normalizers: Comparing and extending network
  normalization schemes.

\bibitem[\protect\astroncite{Salimans et~al.}{2016}]{Salimans-16}
Salimans, T., Goodfellow, I., Zaremba, W., Cheung, V., Radford, A., Chen, X.,
  and Chen, X. (2016).
\newblock Improved techniques for training gans.
\newblock In {\em NIPS}, pages 2234--2242. Curran Associates, Inc.

\bibitem[\protect\astroncite{Salimans and Kingma}{2016}]{Salimans2016WeightNA}
Salimans, T. and Kingma, D.~P. (2016).
\newblock Weight normalization: A simple reparameterization to accelerate
  training of deep neural networks.
\newblock In {\em NIPS}.

\bibitem[\protect\astroncite{Schilling}{2016}]{Schilling-16}
Schilling, F. (2016).
\newblock The effect of batch normalization on deep convolutional neural
  networks.
\newblock Master's thesis, KTH, Centre for Autonomous Systems, CAS.

\bibitem[\protect\astroncite{Schoenholz et~al.}{2016}]{Schoenholz2016DeepIP}
Schoenholz, S.~S., Gilmer, J., Ganguli, S., and Sohl-Dickstein, J. (2016).
\newblock Deep information propagation.
\newblock {\em CoRR}, abs/1611.01232.

\bibitem[\protect\astroncite{Shekhovtsov et~al.}{2018}]{Shekhovtsov-17}
Shekhovtsov, A., Flach, B., and Bu{\v s}ta, M. (2018).
\newblock Feed-forward uncertainty propagation in belief and neural networks.
\newblock {\em CoRR}.

\bibitem[\protect\astroncite{Ulyanov et~al.}{2016}]{Ulyanov-16}
Ulyanov, D., Vedaldi, A., and Lempitsky, V.~S. (2016).
\newblock Instance normalization: The missing ingredient for fast stylization.
\newblock {\em CoRR}, abs/1607.08022.

\bibitem[\protect\astroncite{Wang and Manning}{2013}]{wang2013fast}
Wang, S. and Manning, C. (2013).
\newblock Fast dropout training.
\newblock In {\em ICML}, pages 118--126.

\bibitem[\protect\astroncite{Xiang and Li}{2017}]{xiang2017effects}
Xiang, S. and Li, H. (2017).
\newblock On the effects of batch and weight normalization in generative
  adversarial networks.
\newblock {\em Stat}, 1050:22.

\end{thebibliography}
